\documentclass[letterpaper, 10 pt, conference]{ieeeconf}

\input{glyphtounicode}
\IEEEoverridecommandlockouts
\usepackage{graphicx}
\usepackage{amsmath}
\usepackage{amsfonts}
\usepackage{amssymb}
\usepackage{mathtools}
\usepackage{dsfont}
\usepackage{microtype}
\usepackage{booktabs}
\usepackage{array}
\usepackage{afterpage}
\usepackage{multirow}
\usepackage{url}
\usepackage[T1]{fontenc}
\usepackage[utf8]{inputenc}
\usepackage{xcolor}
\usepackage{arydshln}
\usepackage{hhline}
\usepackage{cite}
\usepackage[bookmarks=true]{hyperref}

\graphicspath{{figs/}}

\centerfigcaptionsfalse

\definecolor{funcolor}{RGB}{150, 78, 66}
\definecolor{graspcolor}{RGB}{122, 96, 78}
\definecolor{zerocolor}{RGB}{67, 94, 128}

\title{\LARGE \bf
{\color{funcolor}Fun}{\color{graspcolor}Grasp}-{\color{zerocolor}Zero}: Field-Conditioned Graph Attention Policy for {\color{zerocolor}Zero}-Shot Cross-Embodiment Dexterous {\color{funcolor}Fun}ctional {\color{graspcolor}Grasp}ing
}
\author{%
\authorblockN{Yanhan Lin$^{*,1}$, Yuliang Wu$^{*,1,2}$, Yihao Guo$^{1}$, Wei-Jin Huang$^{1,2}$, Jianwen Li$^{2}$,\\
Jiawei TANG$^{2}$, Xiaolong Huang$^{2}$, Yuquan Wang$^{2}$, Ancong Wu$^{\dagger,1}$, Wei-Shi Zheng$^{1,3}$}
\authorblockA{$^{1}$Sun Yat-sen University \qquad $^{2}$X SQUARE ROBOT \qquad $^{3}$Shenzhen Loop Area Institute\\
$^{*}$Equal contribution \qquad $^{\dagger}$Corresponding author\\
{\hypersetup{pdfborder={0 0 0}}\href{https://linyh0525.github.io/FunGrasp-Zero/}{\color{magenta}\fontencoding{OT1}\fontfamily{cmtt}\selectfont\urlstyle{same}\nolinkurl{https://linyh0525.github.io/FunGrasp-Zero/}}}}
}

\begin{document}

\maketitle
\thispagestyle{empty}
\pagestyle{empty}

\begin{abstract}
% 需要先讲crossdex研究的必要性
Dexterous functional grasping is crucial for robotic tool use, yet today’s dexterous hands exhibit substantial morphological diversity. This hardware heterogeneity makes cross-embodiment functional grasping a fundamental but largely unexplored problem. 
%定语 functional grasping method不应删
Existing functional grasping methods typically use retargeted hand poses as guidance, which ties the task representation to a specific embodiment and does not explicitly capture object-centric functional contact intent, limiting generalization across hands.
To address this, we propose \textbf{FunGrasp-Zero}, a reinforcement learning framework for zero-shot cross-embodiment functional grasping. 
%改：下面写法突出challenge以及我们的解决思想：接触场表达的其实本质是一种 "intent"，我们引入了物体中心的 手型无关的意图来表达抓取，解决cross embodimentgap
Instead of relying on embodiment-specific joint configuration,  
\textbf{FunGrasp-Zero} captures functional intent with an object-centric, cross-embodiment-consistent contact map, which is adapted to heterogeneous hands via morphological mapping.
%改：下面先引入graph和node概念; 用contact field provides 替代decoupled以保留关联性 
% We represent each hand as a morphology-aligned graph and construct a hand-object potential field from the mapped contact regions, yielding node-wise attractive and repulsive guidance vectors.% for closed-loop control. 
We construct a hand-object potential field from the mapped contact regions, yielding attractive and repulsive guidance vectors. %
% %下面把网络和reward分开细写，增加信息量，比如reward处理contidion的核心目标是什么。基于势能场奖励
A Field-Conditioned Graph Attention Network (FC-GAT) leverages these vectors to coordinate multi-finger motions across hands, while a potential field reward guides fingers toward target regions and penalizes non-target collisions to enforce functional alignment and stability.
We further establish X-FunBench for cross-embodiment functional grasping that evaluates both physical grasp success and functional intent preservation across hands. 
Evaluated on X-FunBench, our unified policy, co-trained on four embodiments, achieves an 86\% average grasp success rate across five unseen hands on novel objects.
Deployed onto four unseen physical hands (Leap, Inspire, Sharpa, and Wuji) to grasp novel objects, the policy achieves an 80\% average success rate without fine-tuning. 
To our knowledge, FunGrasp-Zero is the first RL-based framework to achieve zero-shot cross-embodiment functional grasping across unseen hands in both simulation and the real world.%%claim要强调rl-based

\end{abstract}

\section{Introduction}
\label{sec:introduction}

The rapid growth of dexterous hand platforms has introduced substantial diversity in hand morphology and kinematics, making cross-embodiment generalization an important capability for dexterous manipulation\cite{drograsp,one_policy_fits_all,wei2026one,UHAS,crossdex,trograsp,machagrasp}. 
For robotic tool use, however, functional grasping is not defined by a specific joint configuration, but by task-relevant contacts between particular hand parts and object regions\cite{spahybgen,macontact,secondgrasp,cedex}. 
% Therefore, what should be preserved across embodiments is not a specific joint pose, but the functional contact intent, which each hand can realize through its own morphology and kinematics.
Therefore, cross-embodiment transfer should preserve functional contact intent, which each hand can realize through its own morphology and kinematics rather than specific joint poses.

Existing work has largely studied functional grasping and cross-embodiment dexterous manipulation separately. 
RL-based functional grasping methods achieve closed-loop interaction from human demonstrations,
but typically use retargeted hand poses as embodiment-specific guidance~\cite{codex,fungrasp,demofungrasp}. % Such hand-centric targets couple functional intent with a particular hand configuration, while different hands may require different joint configurations and local contact arrangements to realize the same functional interaction.
Such hand-centric targets tie functional intent to a specific embodiment, although different hands may realize the same interaction differently.
Conversely, recent cross-embodiment frameworks learn transferable representations or control interfaces across different hands, but mainly address generic grasping~\cite{crossdex,dexgraspzero,learning_cross_hand_policies,UHAS,wei2026one,one_policy_fits_all,trograsp}, reaching or in-hand reorientation~\cite{UHAS,getzero} without explicitly modeling functional contact intent. Thus, preserving functional grasp semantics while enabling a unified closed-loop policy to execute them across heterogeneous hands remains an open challenge. To our knowledge, no prior RL-based method has demonstrated zero-shot functional grasping generalization to unseen heterogeneous dexterous hands. %claim一定要强调是closed-loop。静态抓cedex做了

\begin{figure}[!t]
  \centering
  \includegraphics[width=0.9\columnwidth]{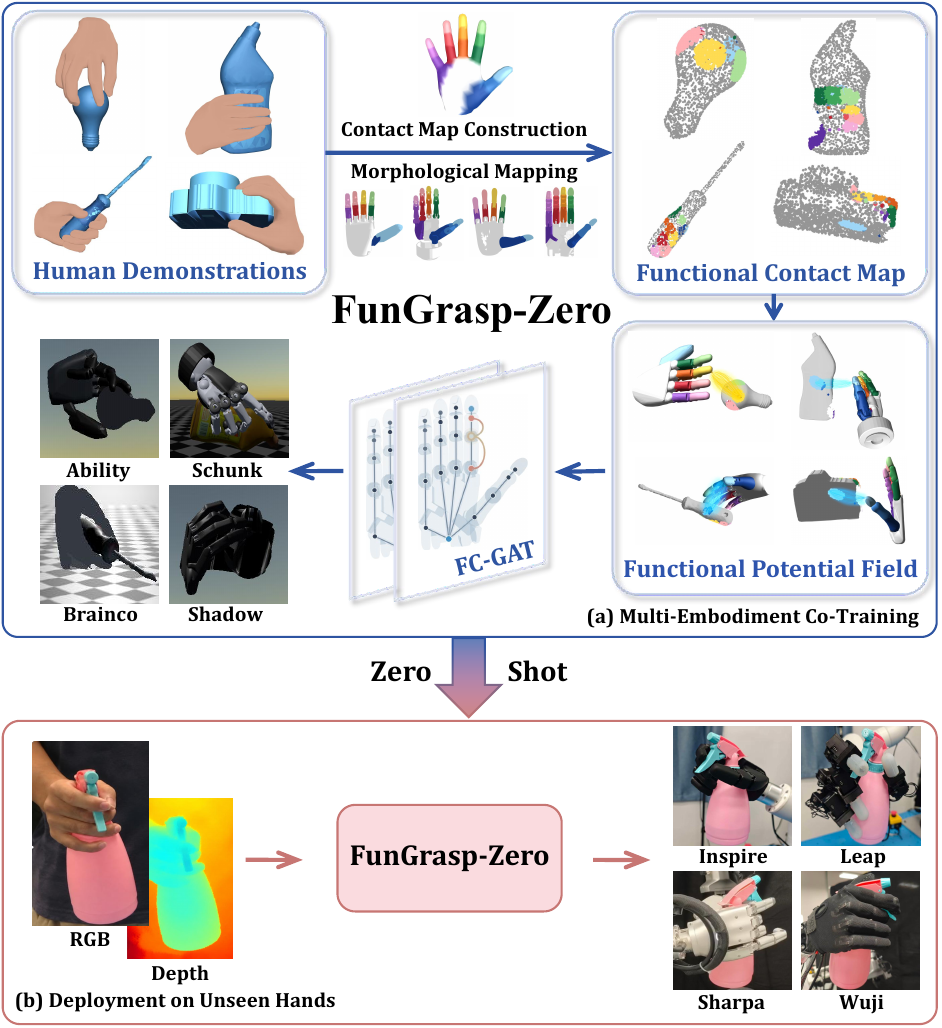}
  % \caption{The X-FunGrasp paradigm. (a) Multi-embodiment co-training: A unified functional grasping policy is trained across diverse hand topologies in simulation using topology-mapped contact fields. (b) Zero-shot transfer: Without embodiment-specific fine-tuning, the frozen policy transfers directly to four unseen physical hands (Inspire, LEAP, Sharpa, Wuji) on novel tools in the real world.}
  \caption{Overview of FunGrasp-Zero. (a) Multi-embodiment co-training of an FC-GAT policy conditioned on functional potential field guidance. (b) Zero-shot deployment on unseen hands and novel tools without fine-tuning.}
  \vspace{-0.5em}
  \label{fig:teaser}
\end{figure}
\afterpage{\begin{figure*}[!t]
  \centering
  \includegraphics[width=\textwidth]{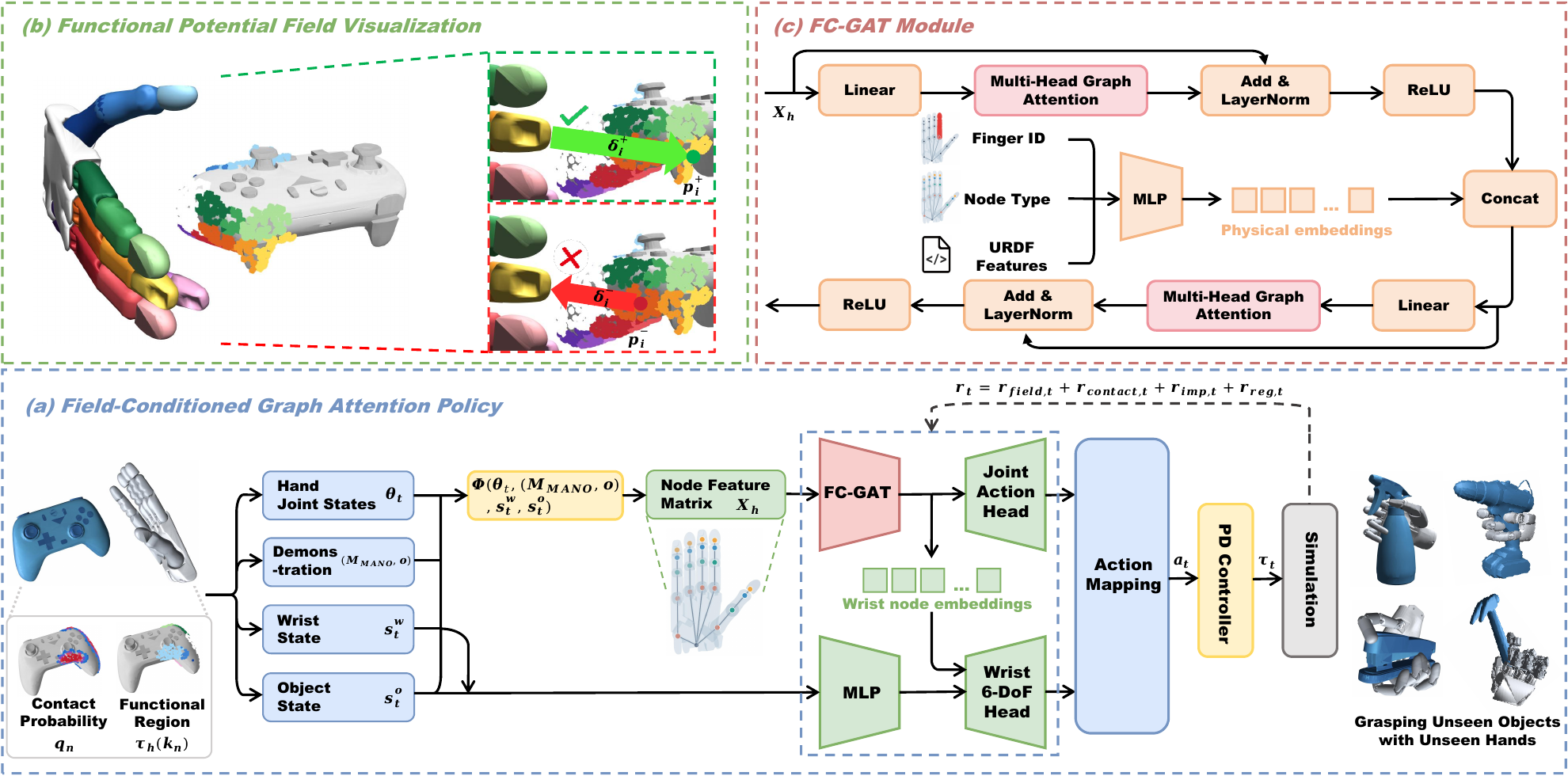}
  % \caption{Overview of the X-FunGrasp framework and policy architecture. (a) Policy pipeline: Hand proprioception, object/wrist states, and the embodiment-specific contact field $\mathcal{C}_h(o)$ are transformed into node features via operator $\Phi$. The CC-GAT policy predicts wrist pose increments and joint motor primitives, which are mapped to physical joint commands for low-level PD execution. (b) Contact-conditioned node features: Spatial attraction/repulsion vectors and contact priors extracted from the contact field relative to target functional regions on the object surface. (c) CC-GAT module: Two cascaded kinematic-masked graph attention blocks conditioned primarily on the contact-conditioned node features from (b) to coordinate multi-finger motions, with URDF physical embeddings concatenated between layers to incorporate hardware boundaries.}
  % \caption{Overview of X-FunGrasp. (a) Policy pipeline: contact fields and robot states are encoded into node features, enabling CC-GAT to predict wrist and joint actions. (b) Contact-conditioned node features capture target attraction, non-target repulsion, and functional contact priors for fine-grained guidance. (c) CC-GAT combines two kinematic-masked graph attention layers with inter-layer URDF physical embeddings for hardware-aware multifingered coordination.}
  \caption{Architecture of FunGrasp-Zero. (a) Field-conditioned graph attention policy: the demonstrated hand--object pair $(\mathcal{M}_{\mathrm{MANO}},o)$ and environment states are processed by the feature extraction operator $\Phi(\cdot)$ to form the node feature matrix $\boldsymbol{X}_h$ for FC-GAT action prediction. (b) Functional potential field $U_o(\boldsymbol{Y})$, where $\boldsymbol{\delta}_i^{+}$ and $\boldsymbol{\delta}_i^{-}$ denote attractive and repulsive guidance for node $i$, with $r_{\mathrm{field}}$ derived from the field. (c) FC-GAT: takes the field-conditioned node feature matrix $\boldsymbol{X}_h$ as input and incorporates morphology-aware physical embeddings.}
  \label{fig:framework}
\end{figure*}
}

To address both limitations, we seek a representation that preserves functional intent while remaining independent of embodiment-specific configurations. An object-centric potential field provides such a representation. We therefore propose \textbf{\textit{FunGrasp-Zero}}, a reinforcement learning framework for zero-shot cross-embodiment functional grasping. 
% 更新势能场概念
Instead of relying on embodiment-specific joint configurations, 
FunGrasp-Zero captures functional intent through an object-centric functional contact map, which is transferred to heterogeneous hands via morphological mapping. 
% Grounded in this mapped contact prior, we construct a hand-object potential field to direct multi-finger reaching. This field provides node-wise spatial guidance vectors and shapes dense rewards, steering designated phalanges toward functional target regions while keeping them away from non-target surfaces. This representation applies directly across heterogeneous hands to accomplish functional grasping tasks.
Grounded in the mapped contact prior, we construct a hand-object potential field that provides node-wise spatial guidance and dense rewards. It attracts designated phalanges toward functional regions while repelling non-target contacts across different hands.

%   A Contact-Conditioned Graph Attention Network (CC-GAT)~\cite{velickovic2018graph} leverages these signals to coordinate multi-finger motions across different hand structures. The same contact field also defines functional interaction rewards, attracting hand parts toward their target regions, penalizing contacts with non-target regions, and regulating contact stability during interaction.
Specifically, from a single human demonstration, we partition the MANO hand~\cite{mano_hand} mesh into phalangeal segments and construct a canonical functional contact map on the object surface. Morphological mapping then transfers these regions to heterogeneous robotic hands according to their kinematic structures. We represent each hand as a morphology-aligned graph. For each hand node, we locate target contact points and non-target repulsive points within the mapped regions to derive spatial guidance vectors, thereby establishing a spatial potential field. Together with physical contact feedback, these guidance vectors form node features. Then a Field-Conditioned Graph Attention Network (FC-GAT) processes them to coordinate multi-finger motions across hands. Meanwhile, this potential field provides dense reinforcement learning rewards, where attractive terms guide fingers toward functional target regions and repulsive terms penalize unintended contacts.

To evaluate zero-shot cross-embodiment functional grasping, we establish \textbf{\textit{X-FunBench}}, which combines diverse functional demonstrations with an evaluation protocol that measures not only grasping success, but also whether functional contact intent is preserved across hands.
On X-FunBench, our unified policy co-trained on four hands achieves a 94\% success rate on seen objects with human-like functional grasps, and maintains 92\% on unseen objects. Crucially, it directly generalizes zero-shot across five heterogeneous hands on novel objects with an 86\% average success rate in simulation. Furthermore, deployed zero-shot onto four physical hands (Leap, Inspire, Wuji, and Sharpa), the policy achieves an 80\% average success rate in the real world.

To the best of our knowledge, we are the first closed-loop RL-based method to achieve zero-shot functional grasping generalization across unseen heterogeneous dexterous hands in both simulation and the real world. Our contributions are summarized as follows:
%\vspace{-0.2em}
\begin{itemize}
    % \item We propose an object-centric functional contact field that represents functional intent independently of embodiment-specific joint configurations and transfers task-relevant contact semantics across heterogeneous hand morphologies through morphological mapping.
    % \item We propose an object-centric functional contact map that transfers task-relevant contact regions across heterogeneous hands via morphological mapping. Grounded in this map, we construct a hand--object potential field that provides node-wise spatial guidance vectors and shapes field-conditioned rewards, driving designated hand parts to establish stable contact on target object surfaces while suppressing unintended contact.
    % \item We propose an object-centric functional contact potential field to capture embodiment-independent functional intent, %which transfers across heterogeneous hand morphologies through morphological mapping.
    \item Based on morphology-aligned graphs, we propose an object-centric functional potential field to capture human grasp intent, which is cross-embodiment consistent. 
    
    \item We design a morphology-aligned graph policy powered by a Field-Conditioned Graph Attention Network (FC-GAT), using potential-field guidance for policy conditioning and reward learning.
    %. By conditioning message passing on node-wise potential field signals and local contact feedback, 
    % , which effectively coordinates multi-finger interactions across diverse kinematic structures.
    
    %We design a morphology-aligned graph policy powered by a Field-Conditioned Graph Attention Network (FC-GAT), using potential-field guidance for policy conditioning and reward learning.

    %\item We establish \textbf{\textit{X-FunBench}}, a benchmark with diverse functional demonstrations and dedicated metrics for evaluating grasp success, functional alignment, and stability across seen and unseen hand embodiments.
    \item We establish X-FunBench, which combines diverse functional demonstrations with dedicated metrics to evaluate grasp success, functional alignment, and stability across seen and unseen hand embodiments.

\end{itemize}
%\vspace{-0.2em}
Our code and benchmark will be made publicly available upon acceptance to contribute to the robotics community.

\section{Related Works}
\label{sec:related_work}

\subsection{Learning-Based Dexterous and Functional Grasping}

% Functional grasping requires contacts on task-relevant object regions that support downstream manipulation while maintaining grasp stability~\cite{codex,simtoolreal}. Learning-based approaches have explored functional grasp synthesis from human demonstrations and object-centric contact information~\cite{wei2023generalized,wu2023functional}, while recent generative methods further extend functional grasp synthesis across hand embodiments using contact-based representations~\cite{cordex}.
Functional grasping targets task-relevant contacts while maintaining grasp stability for downstream manipulation~\cite{codex,simtoolreal}. Learning-based methods have explored functional grasp synthesis from human demonstrations and contact-based representations~\cite{wei2023generalized,wu2023functional,cordex,cedex}, but their open-loop execution is less robust to disturbances. %and does not support zero-shot cross-embodiment generalization. %这句严谨来看你会被纠：cedex是优化方法 按理可以zeroshot，写好优化方程就好。
In parallel, reinforcement learning methods learn closed-loop functional grasping policies from demonstrations~\cite{fungrasp,demofungrasp}, but typically use hand-centric retargeted grasp poses as guidance. This pose-based guidance is embodiment-specific and can lead to contact misalignment across different hand morphologies. Extending closed-loop functional grasping to unseen heterogeneous hand embodiments remains an open challenge. In contrast, our object-centric functional potential field represents functional intent without prescribing a specific hand joint configuration and provides spatial guidance for closed-loop execution across heterogeneous hands.
 %Thus, cross-embodiment functional grasp synthesis has begun to emerge, while closed-loop functional grasp execution across unseen heterogeneous hands remains largely unexplored. %In contrast, FunGrasp-Zero introduces a topology-Aligned Functional Potential Field that provides node-wise functional guidance for closed-loop policy execution across heterogeneous hand morphologies.

\subsection{Cross-Embodiment Dexterous Grasping}
Cross-embodiment dexterous grasping seeks representations and control interfaces that generalize across hands with different morphologies and kinematic structures. Existing methods explore unified action spaces and kinematic mappings~\cite{learning_cross_hand_policies,drograsp,one_policy_fits_all,wei2026one,crossdex,UHAS}, human-like contact representations~\cite{cedex}, morphology-aligned hand graphs~\cite{dexgraspzero}, graph-structured multi-embodiment grasp generation~\cite{graspgraphnet}, and hardware-agnostic interaction representations~\cite{spahybgen,mango_grasp}. These advances improve cross-embodiment grasp generation and control, but most focus on grasp synthesis or generic grasp execution rather than closed-loop task-specific functional interaction. 
FunGrasp-Zero bridges this gap by combining an object-centric functional potential field with a morphology-aligned hand graph, enabling a shared policy to execute consistent functional contacts across unseen heterogeneous hands.
% FunGrasp-Zero addresses this gap by mapping human functional grasp intent into a topology-Aligned Functional Potential Field that provides node-wise guidance for a field-conditioned graph policy, enabling closed-loop functional grasping across heterogeneous hand morphologies.

% \subsection{Cross-Embodiment Representations}
% Cross-embodiment transfer relies on representations that remain compatible across heterogeneous hand morphologies. Existing approaches unify control interfaces through action retargeting~\cite{crossdex}, keypoint kinematic mapping~\cite{learning_cross_hand_policies}, or canonical URDF templates~\cite{wei2026one}. However, these methods often depend on intermediate inverse kinematics or pad varying degrees of freedom with artificial dummy joints. Recent work DexGrasp-Zero~\cite{dexgraspzero} introduces a morphology-aligned hand graph and universal motor primitives, successfully achieving cross-hand generic grasping. While effective for stable object grasping, it does not address human-like functional contact semantics. To this end, we extend cross-embodiment representations to functional interaction tasks, proposing a framework that achieves zero-shot generalization across heterogeneous dexterous hands and novel tools.
% \input{contents/3_preliminaries}
\afterpage{\afterpage{\afterpage{\begin{figure}[!t]
  \centering
  \includegraphics[width=0.8\columnwidth]{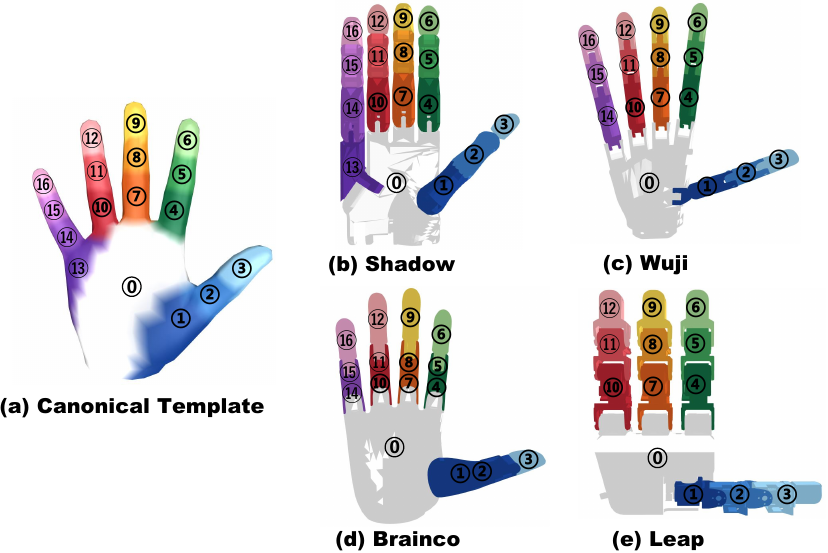}
  \caption{Morphological mapping across hand topologies. The canonical MANO regions in (a) are mapped to robotic hand topologies in (b)–(e). Numbers denote canonical MANO region indices and mapped counterparts. (b) Independent little-finger metacarpal (Shadow), (c) three-phalanx (Wuji), (d) two-phalanx (Brainco), and (e) four-finger (Leap).}
  \vspace{-0.3em}
  \label{fig:morphological_mapping}
\end{figure}
}}}
\afterpage{\afterpage{\afterpage{\afterpage{\afterpage{\afterpage{\begin{figure*}[!t]
  \centering
  \includegraphics[width=\textwidth]{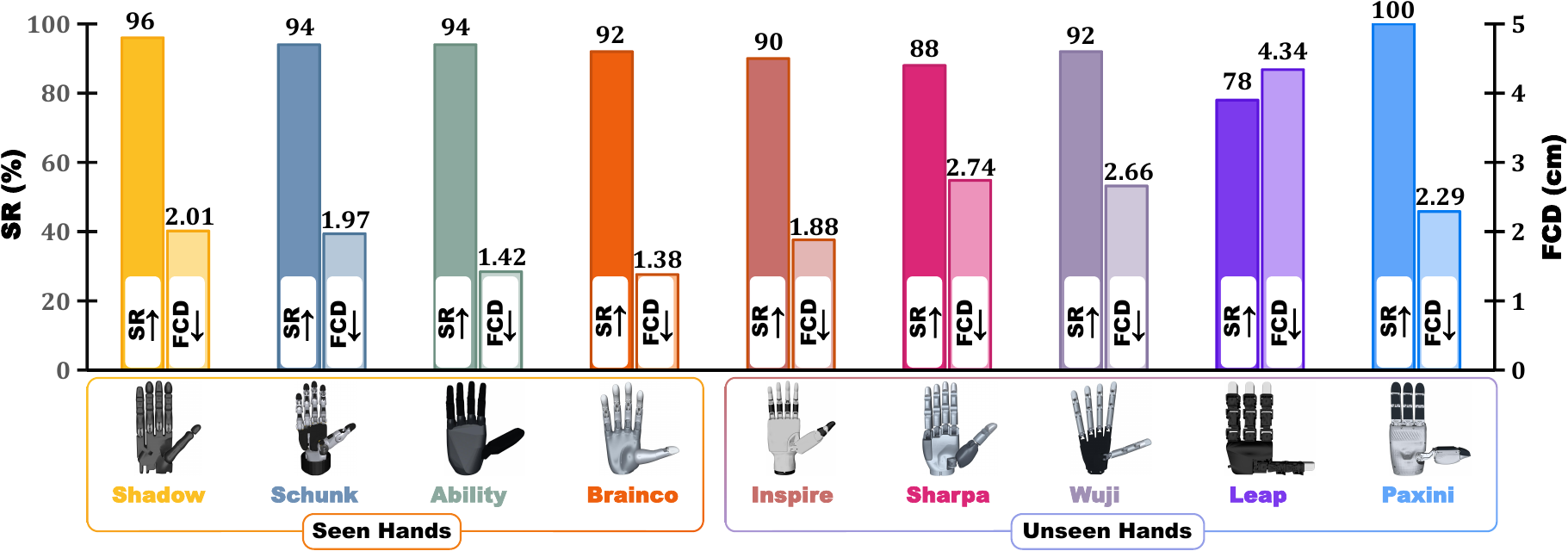}
  \caption{Functional grasping performance across seen and unseen hands on \textit{X-FunBench} training objects. The unified policy is co-trained on four embodiments (Shadow, Schunk, Ability, Brainco) and evaluated zero-shot on five unseen hands (Inspire, Sharpa, Wuji, Leap and Paxini). It achieves a 94\% average success rate and 1.70\,cm contact distance on seen hands, versus 90\% and 2.78\,cm on unseen hands.}
  \label{fig:cross_hand_barchart}
\end{figure*}

}}}}}}

\section{FunGrasp-Zero}
% FunGrasp-Zero is a reinforcement learning framework for zero-shot cross-embodiment dexterous functional grasping, with the overall pipeline depicted in Fig.~\ref{fig:framework}. Given a single human demonstration, it constructs an object-centric functional contact map and projects it onto target robotic hands via morphological mapping. 
% % Each hand is represented as a morphology-aligned graph, where graph nodes correspond to hand joints. Based on this contact map, our framework derives node-wise attractive and repulsive displacement vectors to provide directional spatial guidance as node-level observations, while simultaneously establishing a scalar potential field from these distances to formulate dense functional interaction rewards for policy optimization.
% Each hand is represented as a morphology-aligned kinematic graph, where each graph node corresponds to a joint and is associated with a local functional hand region. 
% Based on the mapped contact map, we construct a hand--object potential field over the graph nodes to provide node-wise attractive and repulsive displacement vectors as directional policy guidance, while its scalar potentials are used to formulate dense functional interaction rewards.
FunGrasp-Zero is a reinforcement learning framework for zero-shot
cross-embodiment dexterous functional grasping. As illustrated in
Fig.~\ref{fig:framework}, each hand is represented as a
morphology-aligned graph. From human demonstrations, we
extract a Topology-Aligned Functional Potential Field to provide
directional guidance and dense rewards for a shared graph-based policy.

\subsection{Problem Formulation}
\label{sec:problem_formulation}

We formulate zero-shot cross-embodiment functional grasping as a goal-conditioned reinforcement learning problem over heterogeneous hand embodiments. During training, hands are sampled from a set of seen embodiments $\mathcal{H}_{\mathrm{train}}$, while zero-shot evaluation is performed on unseen embodiments $\mathcal{H}_{\mathrm{test}}$. %Given a demonstrated hand--object pair $(\mathcal{M}_{\mathrm{MANO}}, o)$, where $\mathcal{M}_{\mathrm{MANO}}$ denotes the MANO hand mesh, we first derive an object-centric functional contact map and transfer it to the target hand morphology to define the functional goal.
Given a demonstrated hand--object pair $(\mathcal{M}_{\mathrm{MANO}},o)$,
we construct an object-centric functional contact map and transfer it to
the target hand via morphological mapping, yielding
$\mathcal{C}_h(o)$. We then derive a hand--object potential field
$U_o(\boldsymbol{Y})$, whose guidance vectors condition the policy and scalar potentials define the field reward.

\indent\textbf{Graph representation and action interface.} Robotic dexterous hands naturally share a tree-like kinematic structure despite substantial differences in kinematics and morphology. Representing a hand as a graph decouples relational kinematic topology from physical dimensions, with nodes as joints and edges as kinematic links, making it inherently suited for cross-embodiment learning. Following DexGrasp-Zero~\cite{dexgraspzero}, we model an arbitrary hand $h$ as a kinematic tree graph $\mathcal{G}_h = (\mathcal{V}_h, \mathcal{E}_h)$ with $N_h = |\mathcal{V}_h|$ nodes, where the adjacency matrix $\boldsymbol{A}^h \in \{0,1\}^{N_h \times N_h}$ captures joint connectivity.

% At control step $t$, the environment state $\boldsymbol{s}_t = (\boldsymbol{\theta}_t, \boldsymbol{s}_t^w, \boldsymbol{s}_t^o)$ comprises the hand joint angles $\boldsymbol{\theta}_t$, wrist state $\boldsymbol{s}_t^w$, and object state $\boldsymbol{s}_t^o$. Conditioned on $\boldsymbol{s}_t$ and the embodiment-specific contact map $\mathcal{C}_h(o)$ derived from $(\mathcal{M}_{\mathrm{MANO}},o)$, the policy $\pi_{\boldsymbol{\phi}}$ predicts an embodiment-agnostic action $\boldsymbol{a}_t$:
% \vspace{-0.4em}
% \begin{equation}
% \boldsymbol{a}_t =
% \left[
% \boldsymbol{a}_{w,t},\, \boldsymbol{a}_{1,t},\, \ldots,\, \boldsymbol{a}_{N_h,t}
% \right]^\top
% \in \mathbb{R}^{6+3N_h},
% \end{equation}
% where $\boldsymbol{a}_{w,t} \in \mathbb{R}^{6}$ represents the 6-DoF wrist displacement, and $\boldsymbol{a}_{i,t} = [\Delta\mathrm{flex}, \Delta\mathrm{abd}, \Delta\mathrm{rot}]^\top \in \mathbb{R}^{3}$ denotes the tri-axial orthogonal motion primitives (flexion--extension, abduction--adduction, and axial rotation) for node $i \in \mathcal{V}_h$. Following DexGrasp-Zero~\cite{dexgraspzero}, an automatic kinematic mapping converts node-wise motion primitives into embodiment-specific joint commands for PD control.
At control step $t$, the environment state $\boldsymbol{s}_t=(\boldsymbol{\theta}_t,\boldsymbol{s}_t^w,\boldsymbol{s}_t^o)$ comprises the hand joint angles, wrist state, and object state. Conditioned on the environment state and field-derived guidance, the field-conditioned policy $\pi_{\boldsymbol{\phi}}$ predicts an embodiment-agnostic action $\boldsymbol{a}_t=[\boldsymbol{a}_{w,t},\boldsymbol{a}_{1,t},\ldots,\boldsymbol{a}_{N_h,t}]^\top\in\mathbb{R}^{6+3N_h}$, where $\boldsymbol{a}_{w,t}\in\mathbb{R}^6$ denotes the wrist pose, and $\boldsymbol{a}_{i,t}\in\mathbb{R}^3$ contains the flexion--extension, abduction--adduction, and axial-rotation primitives for node $i$. These node-wise primitives are automatically mapped to embodiment-specific joint commands for PD control.~\cite{dexgraspzero}

\indent\textbf{Optimization objective.} The policy parameters $\boldsymbol{\phi}$ are optimized to maximize the expected discounted return:
%\vspace{-0.3em}
\begin{equation}
\max_{\boldsymbol{\phi}}
\mathbb{E}_{h \sim \mathcal{H}_{\mathrm{train}},\, o \sim \Omega,\, \boldsymbol{\xi} \sim \pi_{\boldsymbol{\phi}}}
\left[
\sum_{t=0}^{T-1} \gamma^t r_t
\right],
\end{equation}
where $\gamma \in [0,1)$ is the discount factor, and $\boldsymbol{\xi}=[(\boldsymbol{s}_0,\boldsymbol{a}_0),\ldots,(\boldsymbol{s}_{T-1},\boldsymbol{a}_{T-1})]$ denotes a state-action trajectory generated by $\pi_{\boldsymbol{\phi}}$ through interaction with the physics simulation. The reward $r_t$ balances physical grasp stability with functional contact alignment. The resulting policy is then directly evaluated on $h \in \mathcal{H}_{\mathrm{test}}$ without fine-tuning.

\subsection{Topology-Aligned Functional Potential Field}
\label{sec:contact_potential_field}

% Cross-embodiment transfer typically relies on kinematic retargeting, which faces a fundamental trade-off: preserving human-like joint poses causes contact misalignment on the object, while forcing contact alignment yields kinematically infeasible hand poses. Consequently, retargeted poses inherently introduce physical artifacts, making them unsuitable as functional guidance for cross-embodiment learning. %抨击retarget可能口吻不是很好，应该提这不是一个好的cross hand的intent表征。之前方法通常也是用rl在ref pose之上学，kinematically infeasible应该不是一个抨击的点，他们也能在graspose之上学好抓，只是说没有object-centric的指引是很难同时保持手型和接触点。下面细化了表达。
Embodiment-specific hand poses are not a natural representation of cross-embodiment functional intent, since different hands may realize the same task-relevant contacts with different joint configurations. %改成这段 
Therefore, we draw inspiration from artificial potential fields~\cite{khatib1986real} to formulate functional grasping as a contact-matching potential field. Defined over topological functional regions, this field exerts attraction toward intended contacts and repulsion from non-target surfaces, steering heterogeneous hands to fulfill consistent functional intents while respecting their unique kinematics.

\indent\textbf{Canonical contact map construction.} Despite disparities in kinematic and geometric scales, human and robotic hands share a tree topology. As shown in Fig.~\ref{fig:morphological_mapping}(a), we partition the MANO hand mesh $\mathcal{M}_{\text{MANO}}$ into 17 non-overlapping functional regions $\mathcal{F}_{\text{MANO}}$ based on human anatomy: three phalangeal segments per finger and a palm region. An independent metacarpal region for the little finger is included to accommodate hands such as Shadow. Each vertex $\boldsymbol{v} \in \mathcal{M}_{\text{MANO}}$ carries a unique region identifier $f_{\text{label}}(\boldsymbol{v}) \in \mathcal{F}_{\text{MANO}}$.

Given a demonstrated hand--object pair $(\mathcal{M}_{\text{MANO}}, o)$, candidate points $\widetilde{\mathcal{P}}_o = \{\tilde{\boldsymbol{p}}_m\}_{m=1}^M$ are uniformly sampled on the object surface. For each point $\tilde{\boldsymbol{p}}_m$, its non-penetrating distance to the hand mesh is computed as $d_m = \max\left(0, \text{SDF}\left(\tilde{\boldsymbol{p}}_m \mid \mathcal{M}_{\text{MANO}}\right)\right)$, where $\text{SDF}(\cdot)$ denotes the signed distance function. Following Wei et al.~\cite{wei2023generalized}, this distance is converted into a contact probability:
%\vspace{-0.3em}
\begin{equation}
q_m = 2\left(1-\sigma\left(2\alpha d_m\right)\right),
\end{equation}
where $\sigma(\cdot)$ is the sigmoid function and $\alpha$ is a scaling factor.

To filter out irrelevant distant surfaces, we set a contact threshold $\epsilon_c = 0.3$, discarding candidate points with $q_m < \epsilon_c$. From the remaining points, we retain the top $N=1024$ candidates with the highest contact probabilities to form the core set $\mathcal{P}_o = \{\boldsymbol{p}_n\}_{n=1}^N$. Each point $\boldsymbol{p}_n$ is assigned a region label $k_n = f_{\text{label}}(\arg\min_{\boldsymbol{v} \in \mathcal{M}_{\text{MANO}}} \|\boldsymbol{p}_n - \boldsymbol{v}\|_2)$ based on its nearest hand mesh vertex, with candidate points supplemented to ensure every phalange has at least one contact point in $\mathcal{P}_o$. This yields the canonical contact map:
%\vspace{-0.4em}
\begin{equation}
\mathcal{C}_{\text{MANO}}(o) = \left\{ (\boldsymbol{p}_n, q_n, k_n) \mid \boldsymbol{p}_n \in \mathcal{P}_o \right\},
\end{equation}
which partitions $\mathcal{P}_o$ into region-specific contact subsets $\mathcal{P}_o^k$ for each $k \in \mathcal{F}_{\text{MANO}}$.

\indent\textbf{Morphological mapping.} To project the canonical contact map $\mathcal{C}_{\text{MANO}}(o)$ onto diverse robotic morphologies, we define a topological mapping $\tau_h: \mathcal{F}_{\text{MANO}} \to \mathcal{F}_h$ governed by four rules (Fig.~\ref{fig:morphological_mapping}): 1) standard phalanges map one-to-one (e.g., Shadow); 2) hands lacking an independent little-finger metacarpal omit that region (e.g., Wuji); 3) fingers with fewer phalanges merge proximal and intermediate canonical MANO segments into a single robotic link (e.g., Brainco); and 4) four-fingered hands omit the entire little finger (e.g., Leap). Given a hand URDF, $\tau_h$ is automatically instantiated
by parsing its kinematic tree and identifying the applicable rule among the
four predefined topology cases. This yields the embodiment-specific contact map:
%\vspace{-0.3em}
\begin{equation}
\mathcal{C}_h(o) = \left\{ (\boldsymbol{p}_n, q_n, \tau_h(k_n)) \mid \boldsymbol{p}_n \in \mathcal{P}_o \right\}.
\end{equation}
%\vspace{-0.2em}
\indent\textbf{Hand--object potential field.} Let $\boldsymbol{Y} = [\boldsymbol{y}_1, \ldots, \boldsymbol{y}_{N_h}] \in \mathbb{R}^{N_h \times 3}$ denote the 3D positions of all hand nodes in the object frame. Following the artificial potential field formulation~\cite{khatib1986real}, we define the hand--object potential field function $U_o(\boldsymbol{Y})$ as the superposition of node-wise attractive and repulsive potentials, which is minimized to drive the hand toward intended contact regions while avoiding non-target surfaces:
%\vspace{-0.3em}
\begin{equation}
U_o(\boldsymbol{Y}) = \sum_{i \in \mathcal{V}_h} \left( U_{o,i}^+(\boldsymbol{y}_i) + U_{o,i}^-(\boldsymbol{y}_i) \right).
\end{equation}
%\vspace{-0.2em}
For node $i \in \mathcal{V}_h$ with assigned functional region $f_i \in \mathcal{F}_h$, we extract its region-matched subset $\mathcal{P}_o^{f_i} = \{\boldsymbol{p}_n \in \mathcal{P}_o \mid \tau_h(k_n) = f_i\}$ and select the target contact anchor closest to node $i$ from candidates with peak contact probability:
%\vspace{-0.4em}
\begin{subequations}
\begin{gather}
\mathcal{P}_i^+ = \{\boldsymbol{p}_n \in \mathcal{P}_o^{f_i} \mid q_n = \max_{\boldsymbol{p}_j \in \mathcal{P}_o^{f_i}} q_j\}, \\[-1pt]
\boldsymbol{p}_i^+ = \arg\min_{\boldsymbol{p}_n \in \mathcal{P}_i^+} \|\boldsymbol{p}_n - \boldsymbol{y}_i\|_2.
\end{gather}
\end{subequations}%\vspace{-0.2em}
To prevent node $i$ from erroneously contacting object regions intended for other fingers, we collect all points outside target region $f_i$ into the non-target set $\mathcal{P}_i^-$, and select the point closest to node $i$ as the repulsive anchor:
%\vspace{-0.4em}
\begin{subequations}
\begin{gather}
\mathcal{P}_i^- = \{\boldsymbol{p}_n \in \mathcal{P}_o \mid \boldsymbol{p}_n \notin \mathcal{P}_o^{f_i}\}, \\[-1pt]
\boldsymbol{p}_i^- = \arg\min_{\boldsymbol{p}_n \in \mathcal{P}_i^-} \|\boldsymbol{p}_n - \boldsymbol{y}_i\|_2.
\end{gather}
\end{subequations}%\vspace{-0.2em}
Let $q_i^+$ and $q_i^-$ denote the contact probabilities associated with anchors $\boldsymbol{p}_i^+$ and $\boldsymbol{p}_i^-$, respectively.

To eliminate coordinate discrepancies from differing wrist frame conventions across heterogeneous hands, displacement vectors are transformed into a canonical wrist frame:
%\vspace{-0.3em}
\begin{subequations}
\label{eq:vector}
\begin{gather}
\boldsymbol{\delta}_i^+ = (\boldsymbol{R}_w^h)^\top (\boldsymbol{p}_i^+ - \boldsymbol{y}_i), \\[-1pt]
\boldsymbol{\delta}_i^- = (\boldsymbol{R}_w^h)^\top (\boldsymbol{p}_i^- - \boldsymbol{y}_i),
\end{gather}
\end{subequations}%\vspace{-0.2em}
where $\boldsymbol{R}_w^h \in \mathrm{SO}(3)$ is the canonical wrist rotation matrix of hand $h$ in the object frame. The attractive potential $U_{o,i}^+(\boldsymbol{y}_i)$ and repulsive potential $U_{o,i}^-(\boldsymbol{y}_i)$ for node $i$ are formulated as:
%\vspace{-0.3em}
\begin{subequations}
\label{eq:potential_terms}
\begin{gather}
U_{o,i}^+(\boldsymbol{y}_i) = -q_i^+ \exp\left(-\kappa_{\text{att}} \|\boldsymbol{\delta}_i^+\|_2\right), \\[2pt]
U_{o,i}^-(\boldsymbol{y}_i) = q_i^- \exp\left(-\kappa_{\text{rep}} \|\boldsymbol{\delta}_i^-\|_2\right),
\end{gather}
\end{subequations}%\vspace{-0.2em}
where $\kappa_{\text{att}}, \kappa_{\text{rep}}$ are distance scaling factors. Specifically, $U_{o,i}^+$ acts as an attractive term reaching its minimum at target contacts, while $U_{o,i}^-$ serves as a repulsive term peaking at non-target surfaces. With exponential decay, the potential field concentrates steeper gradients near contact surfaces for fine adjustments during close approach. Furthermore, the contact probabilities $q_i^\pm$ weight the field strength, strengthening attraction toward high-probability target regions and repulsion from high-probability non-target regions.

\subsection{Field-Conditioned Graph Attention Policy}
\label{sec:policy_and_reward}

Our policy coordinates multi-finger interactions via FC-GAT, using the guidance vectors $\boldsymbol{\delta}_i^\pm$ from $\mathcal{C}_h(o)$ as input. During training, it optimizes a reward combining the scalar potential field $U_o(\boldsymbol{Y})$ with physical contact feedback.

% \indent\textbf{Node feature representation.} As illustrated in Fig.~\ref{fig:framework}, we construct node-level representations from the environment states and functional priors via a feature extraction operator $\boldsymbol{x}_i = \Phi(\boldsymbol{\theta}_t, \mathcal{C}_h(o), \boldsymbol{s}_t^w, \boldsymbol{s}_t^o)$. For each node $i \in \mathcal{V}_h$, we form feature vector $\boldsymbol{x}_i = [\boldsymbol{\theta}_i, \boldsymbol{\epsilon}_i, \boldsymbol{\delta}_i^+, \boldsymbol{\delta}_i^-, \boldsymbol{u}_i] \in \mathbb{R}^{16}$, where $\boldsymbol{\theta}_i, \boldsymbol{\epsilon}_i \in \mathbb{R}^3$ are the canonical joint angles and PD tracking errors, $\boldsymbol{\delta}_i^+, \boldsymbol{\delta}_i^- \in \mathbb{R}^3$ are the canonical wrist-frame guidance vectors formulated in Section~\ref{sec:contact_potential_field}, and $\boldsymbol{u}_i = [c_i^+, c_i^-, J_i^+, J_i^-] \in \mathbb{R}^4$ captures local physical contact feedback.
\indent\textbf{Node feature representation.} As illustrated in Fig.~\ref{fig:framework}, we obtain node-level features from the demonstration and the current environment state through a feature extraction operator $\boldsymbol{x}_i=\Phi((\mathcal{M}_{\mathrm{MANO}},o),\boldsymbol{\theta}_t,\boldsymbol{s}_t^w,\boldsymbol{s}_t^o)$. For each node $i\in\mathcal{V}_h$, the extracted feature is $\boldsymbol{x}_i=[\boldsymbol{\theta}_i,\boldsymbol{\epsilon}_i,\boldsymbol{\delta}_i^+,\boldsymbol{\delta}_i^-,\boldsymbol{u}_i]\in\mathbb{R}^{16}$, where $\boldsymbol{\theta}_i,\boldsymbol{\epsilon}_i\in\mathbb{R}^3$ denote the canonical joint angles and PD tracking errors, $\boldsymbol{\delta}_i^+,\boldsymbol{\delta}_i^-\in\mathbb{R}^3$ are the canonical wrist-frame guidance vectors derived from the functional potential field in Section~\ref{sec:contact_potential_field}, and $\boldsymbol{u}_i=[c_i^+,c_i^-,J_i^+,J_i^-]\in\mathbb{R}^4$ captures local physical contact feedback.

Specifically, when node $i$ establishes contact with the object surface, it detects a normal impulse $I_i \ge 0$ at contact location $\boldsymbol{p}_{c,i}$. Let $\mathcal{N}(\boldsymbol{p}_{c,i}) = \{ \boldsymbol{p}_n \in \mathcal{P}_o \mid \|\boldsymbol{p}_n - \boldsymbol{p}_{c,i}\|_2 \le r_c \}$ be a local neighborhood of radius $r_c$ centered at the contact point $\boldsymbol{p}_{c,i}$, partitioned into target-matching points $\mathcal{P}_{\mathcal{N},i}^+ = \mathcal{P}_o^{f_i} \cap \mathcal{N}(\boldsymbol{p}_{c,i})$ within node $i$'s designated functional region $\mathcal{P}_o^{f_i}$, and non-matching points $\mathcal{P}_{\mathcal{N},i}^- = \mathcal{P}_i^- \cap \mathcal{N}(\boldsymbol{p}_{c,i})$ from the non-target set $\mathcal{P}_i^-$. If contact occurs within the designated functional region, we record valid contact feedback:
%\vspace{-0.4em}
\begin{subequations}
\begin{gather}
c_i^+ = \max_{\boldsymbol{p}_n \in \mathcal{P}_{\mathcal{N},i}^+} q_n, \quad J_i^+ = I_i, \\
c_i^- = 0, \quad J_i^- = 0;
\end{gather}
\end{subequations}%\vspace{-0.2em}
otherwise, if contact occurs exclusively on non-target surfaces, we assign mismatch feedback:
%\vspace{-0.4em}
\begin{subequations}
\begin{gather}
c_i^- = \max_{\boldsymbol{p}_n \in \mathcal{P}_{\mathcal{N},i}^-} q_n, \quad J_i^- = I_i, \\
c_i^+ = 0, \quad J_i^+ = 0.
\end{gather}
\end{subequations}%\vspace{-0.2em}
In the absence of physical contact, all four components of $\boldsymbol{u}_i$ evaluate to zero. Stacking $\boldsymbol{x}_i$ across all nodes yields the hand feature matrix $\boldsymbol{X}^h \in \mathbb{R}^{N_h \times 16}$.

\indent\textbf{Policy architecture.} As illustrated in Fig.~\ref{fig:framework}, FC-GAT takes the node feature matrix $\boldsymbol{X}^h$ and the kinematic adjacency $\boldsymbol{A}^h$ as input, restricting message passing strictly along the physical kinematic edges $\mathcal{E}_h$. Conditioned on the directional guidance vectors $\boldsymbol{\delta}_i^\pm$ and contact feedback $\boldsymbol{u}_i$ embedded in $\boldsymbol{X}^h$, graph attention weights dynamically modulate each link's dependency on its topological neighbors. Following DexGrasp-Zero~\cite{dexgraspzero}, static morphological attributes are encoded via an MLP into physical embeddings and fused across successive layers to provide hardware-specific physical information during message passing. 

Action generation employs a dual-head MLP decoding architecture: the \textit{Joint Action Head} independently maps each refined link embedding to per-joint motion primitives $\boldsymbol{a}_{i,t} \in \mathbb{R}^3$, while the \textit{Wrist Action Head} concatenates the wrist root node embedding with the external state encoding $(\boldsymbol{s}_t^w, \boldsymbol{s}_t^o)$ to predict the 6-DoF wrist displacement $\boldsymbol{a}_{w,t} \in \mathbb{R}^6$. The assembled action $\boldsymbol{a}_t = [\boldsymbol{a}_{w,t},\, \boldsymbol{a}_{1,t},\, \ldots,\, \boldsymbol{a}_{N_h,t}]^\top$ is then tracked by lower-level PD controllers.

\indent\textbf{Functional reward formulation.} The policy is optimized using a composite reward:
%\vspace{-0.3em}
\begin{equation}
r = r_{\text{field}} + r_{\text{contact}} + r_{\text{imp}} + r_{\text{reg}}.
\end{equation}

The field guidance reward $r_{\text{field}}$ takes the negative of the potential field terms in Eq.~\eqref{eq:potential_terms} to translate potential minimization into reward maximization:
%\vspace{-0.2em}
\begin{equation}
r_{\text{field}} = -\sum_{i \in \mathcal{V}_h} w(i) \left( \lambda_{\text{att}} U_{o,i}^+(\boldsymbol{y}_i) + \lambda_{\text{rep}} D_i^h U_{o,i}^-(\boldsymbol{y}_i) \right),
\end{equation}
where $w(i)$ assigns $w_{\text{tip}}$ to fingertip nodes and $w_{\text{base}}$ to remaining phalanges. Here, $D_i^h$ denotes the shortest topological distance on $\mathcal{G}_h$ between node $i$ and the link associated with anchor $\boldsymbol{p}_i^-$. Weighting the repulsive potential by $D_i^h$ imposes minimal penalty on adjacent links within the same finger while severely penalizing cross-finger intrusions, preventing finger collisions and tangling during closure.

Upon interaction, the contact reward $r_{\text{contact}}$ verifies functional alignment on the object:
%\vspace{-0.3em}
\begin{equation}
r_{\text{contact}} = \sum_{i \in \mathcal{V}_h} w(i) \left( \lambda_{\text{mat}} c_i^+ - \lambda_{\text{mis}} D_i^h c_i^- \right),
\end{equation}
which rewards verified functional contact while penalizing erroneous touches weighted by graph distance $D_i^h$.

To ensure grasp stability, the impulse reward regulates normal interaction forces:
%\vspace{-0.3em}
\begin{equation}
r_{\text{imp}} = \sum_{i \in \mathcal{V}_h} w(i) \left( \lambda_{\text{imp}} \tilde{J}_i^+ - \lambda_{\text{pen}} D_i^h \tilde{J}_i^- \right),
\end{equation}
where $\tilde{J}_i^\pm = \min(J_i^\pm, \bar{J}_i^h)$ clips impulses at $\bar{J}_i^h$ to prevent excessive force exertion. Finally, an action-rate regularization term $r_{\text{reg}} = -\lambda_{\text{reg}} \|\Delta \boldsymbol{\theta}_t^h\|_2^2$ is added to suppress high-frequency joint oscillations.

\section{X-FunBench}
Cross-hand functional grasping requires evaluating not only grasp success, but also whether the same functional intent is preserved across different hand morphologies. Existing benchmarks lack a unified protocol for such cross-hand functional evaluation. We therefore establish \textbf{\textit{X-FunBench}}, which combines functional grasp demonstrations, embodiment-specific contact annotations and cross-hand evaluation metrics.

\textbf{Data.}
X-FunBench provides canonical functional contact annotations that can be mapped to different robotic hands, enabling consistent evaluation across hands. We curate training demonstrations from OakInk~\cite{oakink} and construct the entire test set from newly captured human demonstrations on metric-scale physical tools. Each instance provides the MANO hand pose, metric 6D object pose, and aligned 3D object mesh. 
In total, X-FunBench contains 100 functional grasp demonstrations across 33 object categories and multiple object scales, split into 50 training objects and 50 test objects. 

\textbf{Evaluation Protocol.}
X-FunBench evaluates generalization along both object and hand dimensions. The 50-object test set spans 20 categories, including 8 unseen categories during training and novel object instances from 12 seen categories. For cross-embodiment evaluation, policies are co-trained on four hands (Shadow, Ability, Schunk, and BrainCo) and evaluated zero-shot on five unseen hands (Inspire, Sharpa, Wuji, LEAP, and PaXini). We conduct 20 randomized trials per object-hand pair without embodiment-specific fine-tuning.

\textbf{Evaluation Metrics.}
X-FunBench measures three complementary aspects of functional grasping: \textbf{(1) Success Rate (SR) $\uparrow$:} the percentage of trials in which the object is lifted by more than $0.2\,\mathrm{m}$ and held for at least 20 consecutive time steps. \textbf{(2) Functional Contact Distance (FCD, cm) $\downarrow$:} the average distance from each contacting fingertip to its assigned target functional region. Because target regions are mapped from the same canonical annotations, FCD measures whether functional intent is preserved across heterogeneous hands. \textbf{(3) Relative Wobble Displacement (RWD, mm) $\downarrow$:} the mean inter-step object displacement relative to the wrist frame during lifting, measuring grasp stability.

%\vspace{-0.2em}
\section{Experiments}
%\vspace{-0.3em}
\label{sec:experiments}
We conduct benchmark evaluations on \textit{X-FunBench} in simulation alongside real-world deployments across physical robotic hands, aiming to answer the following five questions: \textbf{Q1:} What zero-shot success rate does our method achieve across unseen hand embodiments and novel objects? \textbf{Q2:} Does the trained policy achieve functional intent alignment with human demonstrations? \textbf{Q3:} How does the single-hand success rate of \textbf{\textit{FunGrasp-Zero}} compare against prior methods relying on kinematic retargeting or fixed grasp patterns? \textbf{Q4:} How do embodiment quantity and morphological diversity influence zero-shot cross-embodiment generalization? \textbf{Q5:} How reliably does our method perform when deployed on physical robotic systems in the real world?
\begin{table}[!t]
\caption{Quantitative comparison of functional grasping performance against baselines on the X-FunBench novel object test set. FCD is in cm and RWD is in mm.}
\label{tab:main_cross_embodiment}
\centering
\footnotesize
\setlength{\tabcolsep}{0pt}
\renewcommand{\arraystretch}{1.2}
\resizebox{\columnwidth}{!}{%
\begin{tabular*}{\columnwidth}{@{\extracolsep{\fill}}
>{\centering\arraybackslash}m{88pt}|
*{3}{>{\centering\arraybackslash}m{27pt}}|
*{3}{>{\centering\arraybackslash}m{27pt}}@{}}
\toprule
\multirow{2}{*}{\textbf{Method}} 
& \multicolumn{3}{c|}{\textbf{Training Hands}} 
& \multicolumn{3}{c}{\textbf{Unseen Hands}} \\
\cmidrule(lr){2-4}
\cmidrule(lr){5-7}
& SR $\uparrow$ & FCD $\downarrow$ & RWD $\downarrow$
& SR $\uparrow$ & FCD $\downarrow$ & RWD $\downarrow$\\
\midrule

DexGrasp-Zero\cite{dexgraspzero}
& 0.85 & 4.86 & 1.43
& 0.78 & 4.62 & 1.86 \\

FunGrasp-Graph\cite{fungrasp}
& 0.76 & 4.52 & 0.48
& 0.43 & 4.38 & 1.46 \\

\hdashline

w/o $\mathcal{C}_h(o)$ \& $U_o(\boldsymbol{Y})$
& 0.87 & 4.68 & 0.39
& 0.71 & 4.81 & 1.12 \\

w/o physical embeddings
& 0.88 & 2.73 & 0.36
& 0.80 & 3.86 & \textbf{0.59} \\

\textbf{Ours}
& \textbf{0.92} & \textbf{2.65} & \textbf{0.27}
& \textbf{0.86} & \textbf{3.60} & 0.61 \\

\bottomrule
\end{tabular*}}
\end{table}
\subsection{Experimental Setup}
\label{sec:exp_setup}
\subsubsection{Simulation Setup}
To evaluate cross-embodiment generalization, we co-train on four dexterous hands spanning diverse physical scales and degrees of freedom, specifically Shadow, Ability, Schunk, and Brainco, and evaluate zero-shot transfer across five unseen hands, including Inspire, Sharpa, Wuji, Leap, and Paxini. In all simulation evaluations, 20 randomized trials are conducted per object to compute average metrics. To measure physical grasp stability alongside functional intent alignment, we follow the X-FunBench evaluation protocol using SR, FCD, and RWD.
\begin{figure}[t]
  \centering
  \includegraphics[width=\linewidth]{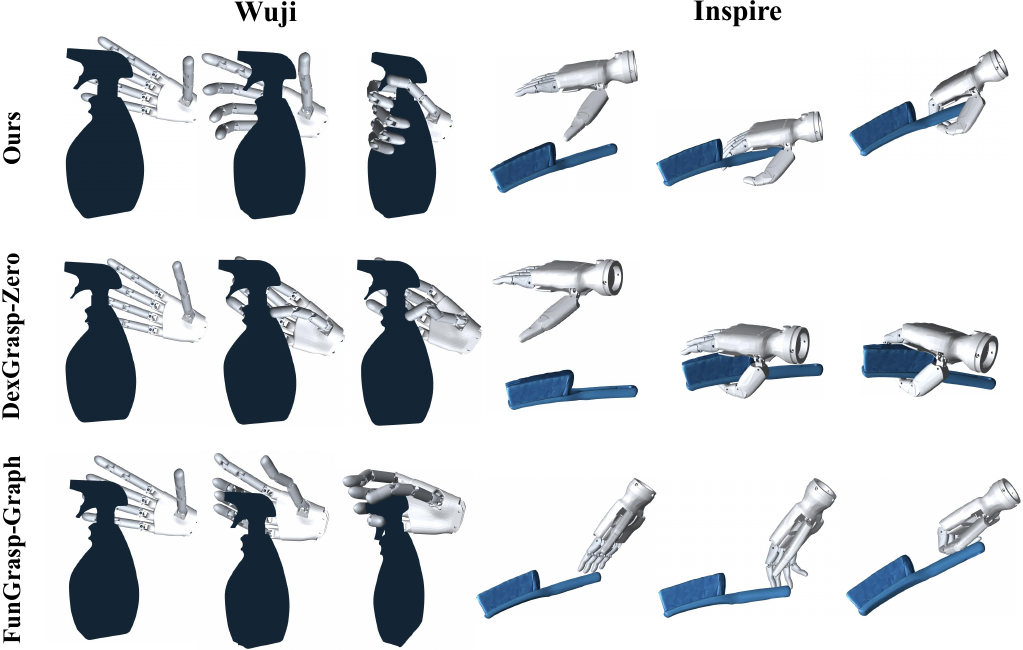}
  \caption{Qualitative comparison of dynamic grasping against cross-embodiment and kinematic retargeting baselines in simulation.}
  % \vspace{-0.4em}
  \label{fig:qualitative_sim}
\end{figure}

\begin{table}[!t]
\caption{Quantitative comparison of functional grasping performance.}
\label{tab:single_hand_comparison}
\centering
\renewcommand{\arraystretch}{1.15}
\scalebox{1.0}{%
\begin{tabular*}{1.00\columnwidth}{@{\extracolsep{\fill}} >{\centering\arraybackslash}m{72pt} | *{3}{>{\centering\arraybackslash}m{45pt}} @{}}
\toprule
\textbf{Method} & SR $\uparrow$ & FCD (cm) $\downarrow$ & RWD (mm) $\downarrow$ \\
\midrule
FunGrasp\cite{fungrasp} & 0.59 & 3.98 & 1.38 \\
DemoFunGrasp\cite{demofungrasp} & 0.72 & 5.16 & 1.51 \\
\textbf{Ours} & \textbf{0.88} & \textbf{2.59} & \textbf{0.53} \\
\bottomrule
\end{tabular*}%
}
\end{table}

\indent\textbf{Baselines}. We compare \textbf{\textit{FunGrasp-Zero}} against representative cross-embodiment and functional grasping methods. DexGrasp-Zero\cite{dexgraspzero} supports cross-embodiment grasping but does not model functional contact intent. 
Since existing functional grasping methods do not directly support our unified multi-hand setting, we construct FunGrasp-Graph as an adapted baseline that represents human functional grasp references using retargeted hand poses following FunGrasp\cite{fungrasp}.
Using the same graph backbone and training setup, it adopts FunGrasp's observations in place of our guidance vectors $(\boldsymbol{\delta}_i^+, \boldsymbol{\delta}_i^-)$ and contact feedback $\boldsymbol{u}_i$, and FunGrasp's pose-tracking and contact-related rewards in place of $r_{\text{field}}$ and $r_{\text{contact}}$, respectively. For single-embodiment comparisons, we additionally evaluate against FunGrasp\cite{fungrasp} and DemoFunGrasp\cite{demofungrasp}, two representative RL-based functional grasping methods guided by hand pose.

\subsubsection{Real-World Evaluation}
We evaluate physical deployment across four unseen hands on two experimental setups: (i) a Kinova Gen3 robotic arm with an Intel RealSense D455 RGB-D camera mounted with either an Inspire hand or a Leap hand; and (ii) a Franka Emika robotic arm with an Orbbec DaBai DCW2 depth camera mounted with a Sharpa or Wuji dexterous hand. Since some of the physical hands, such as the Leap hand, lack tactile sensing, we distill tactile feedback into a unified tactile-free policy following Zhang et al.\cite{robustdexgrasp} for sim-to-real deployment. Benchmark evaluations are performed across 10 novel physical tools spanning multiple geometric scales, with 10 trials conducted per object. For each test object, we first capture a human functional grasp demonstration using the RGB-D camera. We reconstruct the hand-object interaction from the captured RGB and depth data, obtaining the demonstrated hand--object pair $(\mathcal{M}_{\text{MANO}}, o)$ required by our method~\cite{wei2025omnidexgrasp,pwm_artgen}.
% \begin{table}[!t]
% \caption{Zero-shot generalization on novel objects under dual-hand co-training across distinct morphological compositions.}
% \label{tab:dual_hand_comparison}
% \centering
% \scriptsize
% \setlength{\tabcolsep}{2.2pt}
% \renewcommand{\arraystretch}{1.2}

% (a) Morphologically similar high-DoF pair (Shadow + Schunk)\\[2pt]
% \begin{tabular*}{\columnwidth}{@{\extracolsep{\fill}}ccc|cccccc@{}}
% \toprule
% \multicolumn{3}{c|}{\textbf{Training Hands}} & \multicolumn{6}{c}{\textbf{Unseen Hands}} \\
% % \cmidrule(l{2pt}r{0pt}){1-3}\cmidrule(l{0pt}r{2pt}){4-9}
% \cline{1-9}
% Shadow & Schunk & Average & Inspire & Wuji & Sharpa & BrainCo & Ability & Average \\
% \midrule
% 0.88 & 0.79 & \textbf{0.84} & 0.58 & 0.46 & 0.74 & 0.55 & 0.40 & \textbf{0.55} \\
% \bottomrule
% \end{tabular*}

% \vspace{5pt}

% (b) Morphologically diverse pair (Shadow + Ability)\\[2pt]
% \begin{tabular*}{\columnwidth}{@{\extracolsep{\fill}}ccc|cccccc@{}}
% \toprule
% \multicolumn{3}{c|}{\textbf{Training Hands}} & \multicolumn{6}{c}{\textbf{Unseen Hands}} \\
% \cline{1-9}
% Shadow & Ability & Average & Inspire & Wuji & Sharpa & BrainCo & Schunk & Average \\
% \midrule
% 0.82 & 0.90 & \textbf{0.86} & 0.68 & 0.78 & 0.82 & 0.84 & 0.70 & \textbf{0.76} \\
% \bottomrule
% \end{tabular*}
% \end{table}
\begin{table}[!t]
\caption{Zero-shot generalization on novel objects under dual-hand co-training across distinct morphological compositions.}
\label{tab:dual_hand_comparison}
\centering
\scriptsize
\setlength{\tabcolsep}{2.2pt}
\renewcommand{\arraystretch}{1.15}

(a) Morphologically similar high-DoF pair (Shadow + Schunk)\\[3pt]
\begin{tabular*}{\columnwidth}{@{\extracolsep{\fill}}ccc|cccccc@{}}
\toprule
\multicolumn{3}{c|}{\textbf{Training Hands}} & \multicolumn{6}{c}{\textbf{Unseen Hands}} \\
\cmidrule(lr){1-3} \cmidrule(lr){4-9}
Shadow & Schunk & Average & Inspire & Wuji & Sharpa & Brainco & Ability & Average \\
\midrule
0.88 & 0.79 & \textbf{0.84} & 0.58 & 0.46 & 0.74 & 0.55 & 0.40 & \textbf{0.55} \\
\bottomrule
\end{tabular*}

\vspace{8pt}

(b) Morphologically diverse pair (Shadow + Ability)\\[3pt]
\begin{tabular*}{\columnwidth}{@{\extracolsep{\fill}}ccc|cccccc@{}}
\toprule
\multicolumn{3}{c|}{\textbf{Training Hands}} & \multicolumn{6}{c}{\textbf{Unseen Hands}} \\
\cmidrule(lr){1-3} \cmidrule(lr){4-9}
Shadow & Ability & Average & Inspire & Wuji & Sharpa & Brainco & Schunk & Average \\
\midrule
0.82 & 0.90 & \textbf{0.86} & 0.68 & 0.78 & 0.82 & 0.84 & 0.70 & \textbf{0.76} \\
\bottomrule
\end{tabular*}
\end{table}

%\vspace{-0.1em}
\subsection{Implementation Details}
%\vspace{-0.3em}
\label{sec:exp_implementation}
Policies are trained and evaluated using PPO\cite{PPO} in the RaiSim\cite{raisim} simulation with an NVIDIA RTX 3090 GPU. The FC-GAT backbone is implemented as a two-layer GAT. Prior to execution, the initial wrist pose is retracted from the reference demonstration along the palm normal with added spatial perturbations. Each episode comprises a 130-step exploration phase followed by a 50-step lifting phase. Empirically, the spatial decay rates are set to $\kappa_\text{att} = 10.0$ and $\kappa_\text{rep} = 20.0$, node weights to $w_\text{tip} = 3.0$ and $w_\text{base} = 1.0$, and scaling factors to $\lambda_\text{att} = 1.0$, $\lambda_\text{rep} = 0.05$, $\lambda_\text{mat} = 40.0$, $\lambda_\text{mis} = 4.0$, $\lambda_\text{imp} = 16.0$, and $\lambda_\text{pen} = 2.0$.

%\vspace{-0.3em}
\subsection{Cross-Embodiment Functional Grasping}
%\vspace{-0.3em}
% To answer Q1 and Q2, we evaluate per-embodiment performance on \textit{X-FunBench} training objects in Fig.~\ref{fig:cross_hand_barchart}. On training objects, our four-hand co-trained policy achieves a 94\% success rate with an FCD of 1.70\,cm on seen hands, and maintains a 90\% success rate and a 2.78\,cm FCD across five unseen hands. %Even on the four-fingered Leap hand, which is 1.3 times the size of an adult human hand, \textbf{\textit{FunGrasp-Zero}} still secures a 78\% zero-shot success rate, demonstrating strong resilience to substantial morphology and scale differences.
% Even for the large four-fingered Leap hand, \textbf{\textit{FunGrasp-Zero}} achieves a 78\% zero-shot success rate, demonstrating robustness to morphological differences.
To answer Q1 and Q2, we evaluate per-embodiment performance on
\textit{X-FunBench} training objects in Fig.~\ref{fig:cross_hand_barchart}.
Our four-hand co-trained policy achieves 94\% success with a 1.70\,cm
FCD on seen hands, while maintaining 90\% success and 2.78\,cm FCD
across five unseen hands. Even on the large four-fingered Leap hand,
\textbf{\textit{FunGrasp-Zero}} achieves 78\% zero-shot success.
The oversized Leap morphology results in higher FCD, making human-like
functional grasping more difficult, while the grasp success rate
remains high.
%On novel objects as reported in Table~\ref{tab:main_cross_embodiment}, \textbf{\textit{FunGrasp-Zero}} outperforms all baselines across all metrics, reaching an 86\% success rate, a 3.60\,cm FCD, and a small relative wobble displacement of 0.61\,mm on unseen hands. 
On novel objects (Table~\ref{tab:main_cross_embodiment}),
\textbf{\textit{FunGrasp-Zero}} achieves 86\% success, 3.60\,cm FCD,
and 0.61\,mm RWD on unseen hands, outperforming all baselines.
% DexGrasp-Zero maintains strong grasp success but higher FCD indicates generic cross-embodiment grasping does not ensure functional contact alignment. Meanwhile, FunGrasp-Graph drops to a 43\% success rate on unseen hands, %indicating that embodiment-specific retargeted pose guidance transfers poorly across heterogeneous morphologies.
% indicating poor cross-morphology transfer of retargeted pose guidance.
DexGrasp-Zero attains comparable grasp success but higher FCD,
showing that generic cross-embodiment grasping does not guarantee
functional contact alignment. FunGrasp-Graph drops to 43\% success,
indicating poor cross-morphology transfer of retargeted pose guidance.
As visualized in Fig.~\ref{fig:qualitative_sim}, DexGrasp-Zero produces stable but functionally misaligned grasps and FunGrasp-Graph suffers severe contact misalignments, whereas \textbf{\textit{FunGrasp-Zero}} accurately guides designated fingers to exact functional regions such as placing the index finger directly onto a spray trigger.

%\vspace{-0.4em}
\subsection{Single-Hand Training Comparison}
\begin{figure*}[!t]
  \centering
  %\vspace{-0.1em}
  \includegraphics[width=\textwidth]{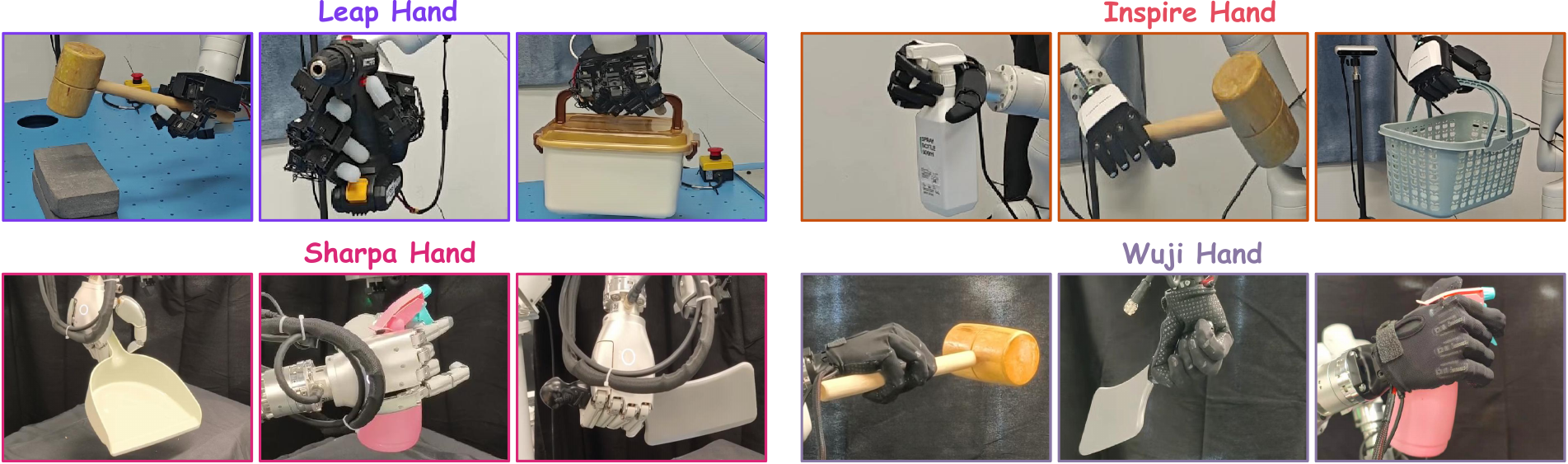}
  \caption{Real-world functional grasping performance across four training-unseen physical hands (Leap, Inspire, Sharpa, and Wuji) on novel physical tools.}
  \label{fig:real_world_performance}
\end{figure*}
\begin{figure}[!t]
  \centering
  \includegraphics[width=0.85\columnwidth]{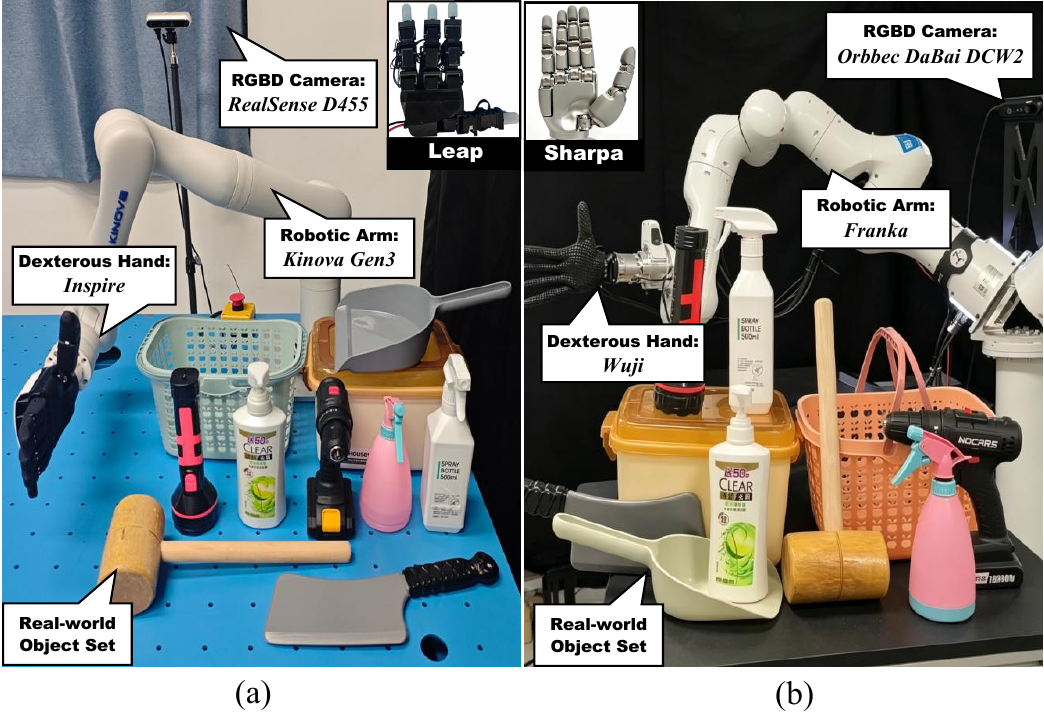}
  %\vspace{-0.1em}
  \caption{Visualization of the real-world hardware platform and objects.}
  %\vspace{-0.1em}
  \label{fig:real_world_hardware}
\end{figure}

\setcounter{table}{3}
% To answer Q3, we evaluate whether the proposed functional representation is also effective in the standard single-embodiment setting. We train dedicated single-hand policies across five embodiments, namely Shadow, Inspire, Sharpa, Ability, and Schunk, on novel objects and benchmark them against specialized single-hand dynamic functional grasping baselines in Table~\ref{tab:single_hand_comparison}. Although FunGrasp\cite{fungrasp} and DemoFunGrasp\cite{demofungrasp} are tailored specifically for single-embodiment training and lack cross-hand capabilities, \textbf{\textit{FunGrasp-Zero}} consistently outperforms both baselines across all hands. This demonstrates that even in single-embodiment settings, our object-centric, topology-aligned functional contact field provides more effective and robust functional interaction guidance than approaches relying on joint-retargeting tracking.
To answer Q3, we evaluate \textbf{\textit{FunGrasp-Zero}} in the single-embodiment setting on novel objects from the \textit{X-FunBench} test set. We train separate policies for five hands, namely Shadow, Inspire, Sharpa, Ability, and Schunk. Table~\ref{tab:single_hand_comparison} reports the average performance across these hands and compares it with specialized single-hand functional grasping baselines. \textbf{\textit{FunGrasp-Zero}} achieves strong performance across the five embodiments, indicating that its functional representation remains effective without relying on cross-embodiment training. 

%\vspace{-0.4em}
\subsection{Impact of Embodiment Diversity and Composition}
To answer Q4, we evaluate how embodiment diversity and quantity affect zero-shot cross-hand generalization. In Table~\ref{tab:dual_hand_comparison}, co-training on two structurally similar high-DoF hands, Shadow and Schunk, yields an average zero-shot success rate of only 0.55 on unseen hands. In contrast, pairing the high-DoF Shadow hand with the low-DoF Ability hand improves the unseen success rate to 0.76. These results suggest that morphological diversity in the training set improves zero-shot transfer by exposing the policy to a broader range of kinematic structures. Building on this observation, our four-hand training setup combines two low-DoF hands with two high-DoF hands, establishing a balanced morphological distribution. As shown in Table~\ref{tab:main_cross_embodiment}, extending to this four-hand configuration raises the unseen hand success rate to 0.86, achieving the best overall performance across both seen and unseen embodiments. %These findings answer Q4, %showing that diverse morphological distribution provides structural coverage while increasing embodiment quantity reinforces multi-finger coordination, jointly driving strong cross-embodiment generalization.
%suggesting that both morphological diversity and the number of training embodiments contribute to stronger cross-embodiment generalization.
The results suggest that both morphological diversity and quantity of training embodiments contribute to stronger cross-embodiment generalization.
% \subsection{Ablation Study}
% \input{contents/table_backbone_ablation}
% We evaluate the core components in Table~\ref{tab:main_cross_embodiment} and assess network architecture design choices in Table~\ref{tab:backbone_ablation}.

% In the component ablation in Table~\ref{tab:main_cross_embodiment}, the variant w/o contact field replaces our functional contact field with nearest-surface point guidance devoid of semantic priors. Its success rate on unseen hands drops to 0.71 and the functional contact distance degrades to 4.81\,cm, confirming that fine-grained contact priors are essential for functional alignment. The variant w/o physical embeddings removes the layer-wise injection of URDF physical attributes, leading to a slight performance decline on unseen hands and demonstrating that hardware-aware physical encoding facilitates cross-embodiment generalization.

% In the network architecture ablation in Table~\ref{tab:backbone_ablation}, Node-MLP fails to coordinate finger motions due to the absence of topological connections. GCN suffers from feature over-smoothing caused by isotropic aggregation, blurring distinct finger roles, while Graph Transformer disrupts the kinematic tree hierarchy due to unconstrained global attention. 

%\vspace{-0.2em}
\subsection{Ablation Study}
%\vspace{-0.2em}
% \begin{table}[!t]
% \caption{Ablation study on policy backbone architectures.}
% \label{tab:backbone_ablation}
% \centering
% \renewcommand{\arraystretch}{1.15}
% \resizebox{\columnwidth}{!}{%
% \begin{tabular}{c | *{3}{>{\centering\arraybackslash}p{45pt}} | *{3}{>{\centering\arraybackslash}p{45pt}}}
% \toprule
% \multirow{2}{*}{\textbf{Method}} 
% & \multicolumn{3}{c|}{\textbf{Training Hands}} 
% & \multicolumn{3}{c}{\textbf{Unseen Hands}} \\
% \cmidrule(lr){2-4} \cmidrule(lr){5-7}
% & SR $\uparrow$ & FCD $\downarrow$ & RWD $\downarrow$ 
% & SR $\uparrow$ & FCD $\downarrow$ & RWD $\downarrow$ \\
% \midrule
% Node-MLP & 0.34 & 3.48 & 1.81 & 0.26 & 3.56 & 1.33 \\
% GCN & 0.78 & 2.55 & \textbf{0.23} & 0.60 & 3.20 & 0.99 \\
% Graph Transformer & 0.69 & 3.15 & 1.38 & 0.64 & 4.54 & 1.82 \\
% \textbf{GAT (Ours)} & \textbf{0.94} & \textbf{1.70} & 0.27 & \textbf{0.90} & \textbf{2.78} & \textbf{0.94} \\
% \bottomrule
% \end{tabular}%
% }
% \end{table}
\begin{table}[!t]
\caption{Ablation study on policy backbone architectures. FCD is in cm and RWD is in mm.}
\label{tab:backbone_ablation}
\centering
\footnotesize
\setlength{\tabcolsep}{0pt}
\renewcommand{\arraystretch}{1.2}
\resizebox{\columnwidth}{!}{%
\begin{tabular*}{\columnwidth}{@{\extracolsep{\fill}}
>{\centering\arraybackslash}m{88pt}|
*{3}{>{\centering\arraybackslash}m{27pt}}|
*{3}{>{\centering\arraybackslash}m{27pt}}@{}}
\toprule
\multirow{2}{*}{\textbf{Method}} 
& \multicolumn{3}{c|}{\textbf{Training Hands}} 
& \multicolumn{3}{c}{\textbf{Unseen Hands}} \\
\cmidrule(lr){2-4}
\cmidrule(lr){5-7}
& SR $\uparrow$ & FCD $\downarrow$ & RWD $\downarrow$
& SR $\uparrow$ & FCD $\downarrow$ & RWD $\downarrow$\\
\midrule

Node-MLP
& 0.34 & 3.48 & 1.81
& 0.26 & 3.56 & 1.33 \\

GCN
& 0.78 & 2.55 & \textbf{0.23}
& 0.60 & 3.20 & 0.99 \\

Graph Transformer
& 0.69 & 3.15 & 1.38
& 0.64 & 4.54 & 1.82 \\

\hdashline

\textbf{GAT (Ours)}
& \textbf{0.94} & \textbf{1.70} & 0.27
& \textbf{0.90} & \textbf{2.78} & \textbf{0.94} \\

\bottomrule
\end{tabular*}}
\end{table}
\begin{table}[!t]
\caption{Real-world functional grasping success rates on new objects.}
\label{tab:real_world_results}
\centering
\footnotesize
\setlength{\tabcolsep}{4.0pt}
\renewcommand{\arraystretch}{1.2}
\begin{tabular*}{\columnwidth}{@{\extracolsep{\fill}}c|ccccc@{}}
\toprule
\textbf{Method} & \textbf{Leap} & \textbf{Inspire} & \textbf{Sharpa} & \textbf{Wuji} & \textbf{Average} \\
\midrule
Single-hand & 0.81 & 0.84 & 0.90 & 0.86 & 0.85 \\
Cross-hand & 0.76 & 0.82 & 0.80 & 0.83 & 0.80 \\
\bottomrule
\end{tabular*}
\end{table}

% In the component ablation (Table~\ref{tab:main_cross_embodiment}), the variant \textit{w/o} $\mathcal{C}_h(o)$ \& $U_o(\boldsymbol{Y})$ strips both the functional contact map and the potential field. Specifically, it eliminates the field guidance reward $r_{\text{field}}$, while replacing the region-conditioned guidance vectors $(\boldsymbol{\delta}_i^\pm)$ and contact feedback $\boldsymbol{u}_i$ with Euclidean distances to the nearest object surface points. %Its success rate on unseen hands drops to 0.71 and functional contact distance degrades to 4.81\,cm, confirming that coupling topological contact maps with potential field shaping is critical for functional intent alignment. The variant \textit{w/o} physical embeddings omits the layer-wise injection of URDF attributes, leading to degraded generalization on unseen embodiments and demonstrating that hardware-aware physical encoding facilitates cross-embodiment transfer.
In the component ablation (Table~\ref{tab:main_cross_embodiment}), the
\textit{w/o} $\mathcal{C}_h(o)$ \& $U_o(\boldsymbol{Y})$ variant replaces
the contact feedback $\boldsymbol{u}_i$ and reward $r_{\text{contact}}$
with generic contact signals, without distinguishing functional-region
contacts. It also removes $r_{\text{field}}$ and replaces the
region-conditioned guidance vectors $(\boldsymbol{\delta}_i^\pm)$ with
nearest-surface Euclidean distances.
Its SR on unseen hands drops to 0.71 and FCD degrades to 4.81\,cm, indicating that the proposed functional potential field and the derived guidance are important for functional grasping and cross-embodiment transfer. The \textit{w/o} physical embeddings variant, which removes the
inter-layer physical embeddings in FC-GAT, also degrades unseen-hand
performance, suggesting that hardware-specific attributes benefit
cross-embodiment transfer.

We further compare different backbones on training objects in
Table~\ref{tab:backbone_ablation}. Node-MLP ignores hand kinematic
topology, limiting multi-finger coordination, while GCN relies on fixed
neighbor aggregation. Graph Transformer models broader node interactions
without explicitly preserving local kinematic structure. In contrast,
GAT adaptively aggregates features along kinematic edges, enabling
reliable grasping, precise functional contact alignment, and stable
object handling.

% \vspace{-0.3em}
% \subsection{Real-World Experiments}
% \input{contents/table_real_world_results}
% In physical deployments, we use FoundationPose\cite{foundationpose} to estimate 6D poses of novel objects. As shown in Table~\ref{tab:real_world_results}, when directly deploying the unified four-hand co-trained policy zero-shot onto four unseen physical hands on novel tools, the average success rate is close to that of dedicated single-hand policies. This result verifies that our method possesses robust sim-to-real transfer capabilities and can reliably generalize to unseen physical hardware platforms for functional grasping.

%\vspace{-0.4em}
\subsection{Real-World Experiments}
As shown in Table~\ref{tab:real_world_results}, the unified four-hand co-trained policy achieves an 80\% average success rate across four unseen physical hands on novel tools, compared with 85\% for dedicated single-hand policies. %, demonstrating effective zero-shot sim-to-real transfer.
A trial is successful only when all three independent evaluators agree that the grasp preserves the demonstrated functional intent, demonstrating effective zero-shot sim-to-real transfer.
%\vspace{-0.4em}
\section{Conclusion}
\label{sec:conclusion}

In this paper, we proposed \textbf{\textit{FunGrasp-Zero}}, a reinforcement learning framework for zero-shot cross-embodiment dexterous functional grasping. 
% By transferring an object-centric functional contact map across different hands and deriving a hand--object potential field for policy guidance and reward shaping, FunGrasp-Zero enables FC-GAT to coordinate functional multi-finger interactions across embodiments.
% By adapting an object-centric functional contact field to heterogeneous hands via morphological mapping, \textit{FunGrasp-Zero} extracts contact-conditioned node features and functional rewards to guide FC-GAT in coordinating multi-finger motions. 
By constructing a functional potential field across different hands, FunGrasp-Zero uses field-conditioned node features and functional rewards to guide FC-GAT in coordinating multi-finger motions. 
Evaluations on X-FunBench and zero-shot deployments across four unseen physical hands demonstrate robust generalization and functional intent alignment across unseen hands and objects.
\vspace{-0.3em}

\bibliographystyle{IEEEtran}
\bibliography{references}

@String{CVPR = {CVPR}}

@String{ECCV = {ECCV}}

@String{ICLR = {ICLR}}

@String{RSS = {RSS}}

@String{CoRL = {CoRL}}

@String{ICRA = {ICRA}}

@inproceedings{oakink,
  author = {Yang, Lixin and Li, Kailin and Zhan, Xinyu and Wu, Fei and Xu, Anran and Liu, Liu and Lu, Cewu},
  title = {{OakInk}: A Large-Scale Knowledge Repository for Understanding Hand-Object Interaction},
  booktitle = {CVPR},
  year = {2022}
}

@inproceedings{trograsp,
  title = {T(R,O) Grasp: Efficient Graph Diffusion of Robot-Object Spatial Transformation for Cross-Embodiment Dexterous Grasping},
  author = {Xin Fei and Zhixuan Xu and Huaicong Fang and Tianrui Zhang and Lin Shao},
  booktitle = {ICRA},
  year = {2026}
}

@article{secondgrasp,
  title = {SECOND-Grasp: Semantic Contact-guided Dexterous Grasping},
  author = {Han Yi Shin and Heeju Ko and Jaewon Mun and Qixing Huang and Jaehyeok Lee and Sung June Kim and Honglak Lee and Sujin Jang and Sangpil Kim},
  journal = {arXiv:2605.13117},
  year = {2026}
}

@inproceedings{macontact,
  author = {Ma, Yiyao and Chen, Kai and Zheng, Kexin and Dou, Qi},
  title = {Contact Map Transfer with Conditional Diffusion Model for Generalizable Dexterous Grasp Generation},
  booktitle = {NeurIPS},
  year = {2025}
}

@article{raisim,
  title = {Per-contact iteration method for solving contact dynamics},
  author = {Hwangbo, Jemin and Lee, Joonho and Hutter, Marco},
  journal = {IEEE Robotics and Automation Letters},
  year = {2018}
}

@article{khatib1986real,
  title = {Real-time obstacle avoidance for manipulators and mobile robots},
  author = {Khatib, Oussama},
  journal = {The International Journal of Robotics Research},
  year = {1986}
}

@inproceedings{robustdexgrasp,
  author = {Hui Zhang and Zijian Wu and Linyi Huang and Sammy Christen and Jie Song},
  title = {RobustDexGrasp: Robust Dexterous Grasping of General Objects},
  booktitle = {CoRL},
  year = {2025}
}

@inproceedings{crossdex,
  author = {Yuan, Haoqi and Zhou, Bohan and Fu, Yuhui and Lu, Zongqing},
  title = {Cross-Embodiment Dexterous Grasping with Reinforcement Learning},
  booktitle = {ICLR},
  year = {2025}
}

@inproceedings{dexgraspzero,
  title = {DexGrasp-Zero: A Morphology-Aligned Policy for Zero-Shot Cross-Embodiment Dexterous Grasping},
  author = {Yuliang Wu and Yanhan Lin and WengKit Lao and Yuhao Lin and Yi-Lin Wei and Wei-Shi Zheng and Ancong Wu},
  booktitle = {RSS},
  year = {2026}
}

@inproceedings{simtoolreal,
  title = {SimToolReal: An Object-Centric Policy for Zero-Shot Dexterous Tool Manipulation},
  author = {Kushal Kedia and Tyler Ga Wei Lum and Jeannette Bohg and C. Karen Liu},
  booktitle = {RSS},
  year = {2026}
}

@article{wei2023generalized,
  author = {Wei, Wei and Wang, Peng and Wang, Sizhe and Luo, Yongkang and Li, Wanyi and Li, Daheng and Huang, Yayu and Duan, Haonan},
  title = {Learning Human-Like Functional Grasping for Multifinger Hands From Few Demonstrations},
  journal = {IEEE Transactions on Robotics},
  year = {2024}
}

@article{wu2023functional,
  author = {Wu, Rina and Zhu, Tianqiang and Peng, Wanli and Hang, Jinglue and Sun, Yi},
  title = {Functional Grasp Transfer Across a Category of Objects From Only one Labeled Instance},
  journal = {IEEE Robotics and Automation Letters},
  year = {2023}
}

@article{UHAS,
  title = {Cross-Embodiment Robot Manipulation via a Unified Hand Action Space},
  author = {Luis Felipe Casas and Robert Teal and Keval Shah and Abhijit Tadepalli and Wan-Xin Jin and Yu Xiang},
  journal = {arXiv:2607.03570},
  year = {2026}
}

@inproceedings{codex,
  author = {Bowen Jiang and William Painter Reger and Roberto Martin-Martin},
  title = {CoDex: Learning Compositional Dexterous Functional Manipulation without Demonstrations},
  booktitle = {ICRA},
  year = {2026}
}

@inproceedings{machagrasp,
  title = {MachaGrasp: Morphology-Aware Cross-Embodiment Dexterous Hand Articulation Generation for Grasping},
  author = {Heng Zhang and Kevin Yuchen Ma and Mike Zheng Shou and Weisi Lin and Yan Wu},
  booktitle = {ICRA},
  year = {2026}
}

@article{fungrasp,
  author = {Huang, Linyi and Zhang, Hui and Wu, Zijian and Christen, Sammy and Song, Jie},
  title = {FunGrasp: Functional Grasping for Diverse Dexterous Hands},
  journal = {IEEE Robotics and Automation Letters},
  year = {2025}
}

@inproceedings{demofungrasp,
  title = {DemoFunGrasp: Universal Dexterous Functional Grasping via Demonstration-Editing Reinforcement Learning},
  author = {Chuan Mao and Haoqi Yuan and Ziye Huang and Chaoyi Xu and Kai Ma and Zongqing Lu},
  booktitle = {CVPR},
  year = {2026}
}

@inproceedings{drograsp,
  author = {Wei, Zhenyu and Xu, Zhixuan and Guo, Jingxiang and Hou, Yiwen and Gao, Chongkai and Cai, Zhehao and Luo, Jiayu and Shao, Lin},
  title = {$\mathcal{D}(\mathcal{R}, \mathcal{O})$ Grasp: A Unified Representation of Robot and Object Interaction for Cross-Embodiment Dexterous Grasping},
  booktitle = {ICRA},
  year = {2025}
}

@inproceedings{learning_cross_hand_policies,
  title = {Learning Cross-Hand Policies of High-DOF Reaching and Grasping},
  author = {Qijin She and Shishun Zhang and Yunfan Ye and Ruizhen Hu and Kai Xu},
  booktitle = {ECCV},
  year = {2024}
}

@inproceedings{wei2026one,
  title = {One Hand to Rule Them All: Canonical Representations for Unified Dexterous Manipulation},
  author = {Wei, Zhenyu and Yao, Yunchao and Ding, Mingyu},
  booktitle = {RSS},
  year = {2026}
}

@inproceedings{one_policy_fits_all,
  author = {Juncheng Mu and Sizhe Yang and Hojin Bae and Feiyu Jia and Qingwei Ben and Boyi Li and Huazhe Xu and Jiangmiao Pang},
  title = {One-Policy-Fits-All: Geometry-Aware Action Latents for Cross-Embodiment Manipulation},
  booktitle = {ICRA},
  year = {2026}
}

@inproceedings{wei2025omnidexgrasp,
  title = {OmniDexGrasp: Generalizable Dexterous Grasping via Foundation Model and Force Feedback},
  author = {Yi-Lin Wei and Zhexi Luo and Yuhao Lin and Mu Lin and Zhizhao Liang and Shuoyu Chen and Wei-Shi Zheng},
  booktitle = {ICRA},
  year = {2026}
}

@inproceedings{cedex,
  author = {Wu, Zhiyuan and Potamias, Rolandos Alexandros and Zhang, Xuyang and Zhang, Zhongqun and Deng, Jiankang and Luo, Shan},
  title = {CEDex: Cross-Embodiment Dexterous Grasp Generation at Scale from Human-Like Contact Representations},
  booktitle = {ICRA},
  year = {2026}
}

@inproceedings{cordex,
  author = {He, Xingyi and Polavaram, Adhitya and Cao, Yunhao and Deshmukh, Om and Wang, Tianrui and Zhou, Xiaowei and Fang, Kuan},
  title = {Generate, Transfer, Adapt: Learning Functional Dexterous Grasping from a Single Human Demonstration},
  booktitle = {ICRA},
  year = {2026}
}

@article{graspgraphnet,
  author = {Yeonseo Lee and Taeyeop Lee and Hyosup Shin and Guebin Hwang and Sungho Jo},
  title = {GraspGraphNet: Graph-Structured Multi-Embodiment Dexterous Grasp Generation},
  journal = {arXiv:2607.11031},
  year = {2026}
}

@article{mango_grasp,
  title = {MANGO-Grasp: Mahalanobis Fields over Geometry-Oriented 3D Gaussians for Cross-Embodiment Dexterous Grasping},
  author = {Heng Zhang and Kevin Yuchen Ma and Mike Zheng Shou and Weisi Lin and Yan Wu},
  journal = {arXiv:2608.02014},
  year = {2026}
}

@article{spahybgen,
  author = {Xianli Wang and Lap Mou Tam and Qingsong Xu},
  title = {Learning Contact Representations in Real-World Clutter for Universal Robotic Grasping},
  journal = {Nature Machine Intelligence},
  year = {2026}
}

@inproceedings{getzero,
  title = {{GET-Zero}: Graph Embodiment Transformer for Zero-shot Embodiment Generalization},
  author = {Austin Patel and Shuran Song},
  booktitle = {ICRA},
  year = {2025}
}

@article{mano_hand,
  title = {Embodied hands: modeling and capturing hands and bodies together},
  author = {Romero, Javier and Tzionas, Dimitrios and Black, Michael J.},
  journal = {ACM Transactions on Graphics},
  year = {2017}
}

@article{PPO,
  author = {John Schulman and Filip Wolski and Prafulla Dhariwal and Alec Radford and Oleg Klimov},
  title = {Proximal Policy Optimization Algorithms},
  journal = {arXiv:1707.06347},
  year = {2017}
}

@inproceedings{pwm_artgen,
  title = {{PWM-ArtGen}: Part World Model for Articulated Object Generation},
  author = {Zheng, Wentao and Wu, Ancong},
  booktitle = {ECCV},
  year = {2026},
}

\end{document}